\documentclass[10pt,twocolumn,letterpaper]{article}

\usepackage{iccv}
\usepackage{authblk}
\usepackage{times}
\usepackage{epsfig}
\usepackage{graphicx}
\usepackage{amsmath}
\usepackage{amssymb}
\usepackage{enumitem}
\usepackage{multirow}
\usepackage{booktabs}
\usepackage{cite}
\usepackage[T1]{fontenc}
\usepackage[utf8]{inputenc}
\usepackage[english]{babel}
\usepackage[autostyle, english=american]{csquotes}
\MakeOuterQuote{"}

\usepackage[breaklinks=true,bookmarks=false]{hyperref}

\iccvfinalcopy % *** Uncomment this line for the final submission

\def\iccvPaperID{****} % *** Enter the ICCV Paper ID here

\ificcvfinal\fi

\begin{document}

%%%%%%%%% TITLE
\title{The Solution for Temporal Action Localisation Task of  Perception Test Challenge 2024 }

%\author{Dian Chao\\
%Nanjing University of Science and Technology\\
%\and
%%Shupeng Zhong\\
%Nanjing University of Science and Technology\\

\author{
  Yinan Han$^1$,
  Qingyuan Jiang$^1$,
  Hongming Mei$^2$,
 % Yang Yang$^1$,
  Yang Yang\textsuperscript{1}\thanks{Corresponding Author},
  Jinhui Tang$^1$
}

\affil{
  $^1$Nanjing University of Science and Technology\\
  $^2$University of Toronto\\
  
}

\maketitle
\ificcvfinal\thispagestyle{empty}\fi

\begin{abstract}
\quad This report presents our method for Temporal Action Localisation (TAL), which focuses on identifying and classifying actions within specific time intervals throughout a video sequence. We employ a data augmentation technique by expanding the training dataset using overlapping labels from the Something-SomethingV2 dataset, enhancing the model’s ability to generalize across various action classes. For feature extraction, we utilize state-of-the-art models, including UMT, VideoMAEv2 for video features, and BEATs and CAV-MAE for audio features. Our approach involves training both multimodal (video and audio) and unimodal (video only) models, followed by combining their predictions using the Weighted Box Fusion (WBF) method. This fusion strategy ensures robust action localisation. our overall approach achieves a score of 0.5498, securing first place in the competition.
\end{abstract}

\maketitle

\section{Introduction}
In recent years, deep learning techniques have gained significant attention across numerous research fields~\cite{YangWZL018,YangHGXX23,YangYBZZGXY23,YangWZX019}. The Perception Test Temporal Action Localisation Challenge 2024 is part of the broader Perception Test Challenge, aiming to comprehensively evaluate the perception and reasoning capabilities of multimodal video models~\cite{YangFZLJ21}. Unlike traditional benchmarks that focus primarily on computational performance, this challenge emphasizes perceptual tasks by leveraging real-world video data that are intentionally designed, filmed, and annotated. The goal is to assess a wide range of skills, reasoning types, and modalities within multimodal perception models.

\begin{figure}[h]
 \centering
 \includegraphics[width=0.5\textwidth]{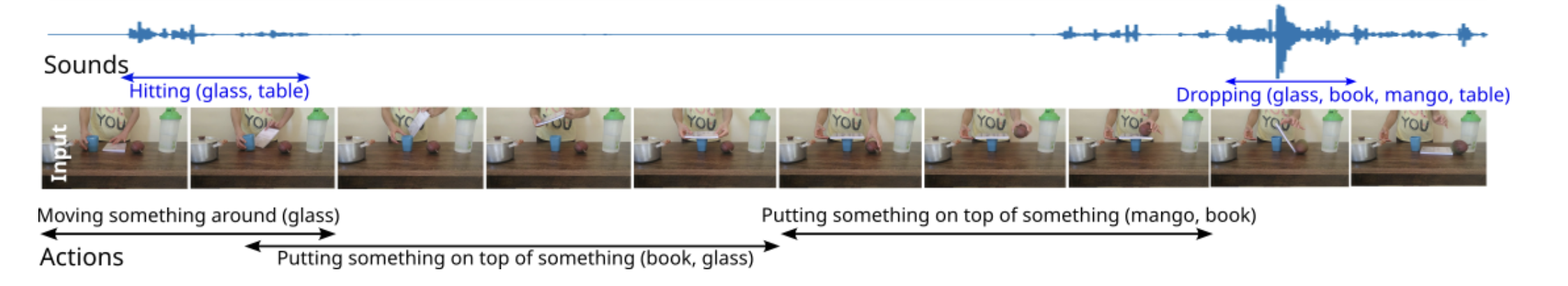}
 \caption{the examples of Action Localisation annotations}
 \label{fig1}
\end{figure}

In the Temporal Action Localis
ation (TAL) task, the objective is to develop a method that can accurately localize and classify actions occurring within untrimmed videos according to a predefined set of classes. Each action is represented by start and end timestamps along with its corresponding class label, as illustrated in Figure\ref{fig1}. This task is critical for various applications, including video surveillance, content analysis, and human-computer interaction.The dataset provided for this challenge is derived from the Perception Test, comprising high-resolution videos (up to 35 seconds long, 30fps, and a maximum resolution of 1080p). Each video contains multiple action segment annotations. To facilitate experimentation, both video and audio features are provided, along with detailed annotations for the training and validation phases. 

To tackle this challenge, we experimented with various models such as UMT~\cite{DBLP:conf/cvpr/LiuLWCSQ22}, InternalVideo~\cite{DBLP:journals/corr/abs-2403-15377}, and VideoMAE~\cite{DBLP:conf/nips/TongS0022}. Ultimately, we selected UMT~\cite{DBLP:conf/cvpr/LiuLWCSQ22} and VideoMAEv2~\cite{DBLP:conf/cvpr/WangHZTHWWQ23} as our video feature extraction models due to their superior ability to capture temporal dynamics in video sequences. For audio feature extraction, we employed BEATs~\cite{DBLP:conf/icml/ChenW00T0CYW23} and CAV-MAE~\cite{DBLP:conf/iclr/GongRLHKKG23}, which effectively capture essential audio patterns and integrate audio-visual information. Additionally, we augmented our training dataset by incorporating overlapping labels from the Something-SomethingV2 dataset—a large-scale, crowd-sourced video dataset focusing on basic actions and interactions with everyday objects. By integrating these overlapping action classes, we enriched our dataset and enhanced the model's generalization capabilities across different scenarios. Finally, we fused the predictions from our models using the Weighted Box Fusion (WBF)~\cite{DBLP:journals/ivc/SolovyevWG21} method, which combines the strengths of each model's outputs. Through this comprehensive approach, we achieved first place in the competition.

\begin{figure*}[h]
 \centering
 \includegraphics[width=\textwidth]{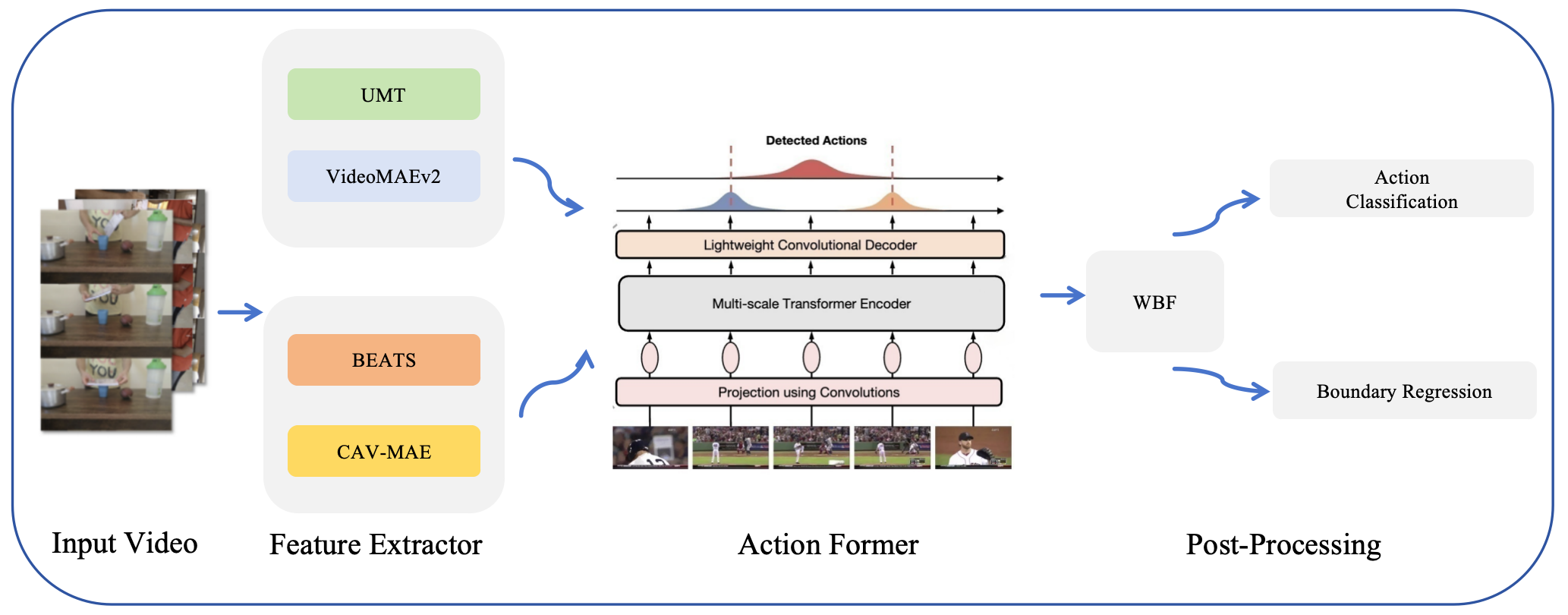}
 \caption{Flowchart of our proposed method. The process begins with the Video Input and Audio Inpu, which are separately processed to extract features using pretrained models (UMT~\cite{DBLP:conf/cvpr/LiuLWCSQ22}, VideoMAEv2~\cite{DBLP:conf/cvpr/WangHZTHWWQ23} for video; BEATs~\cite{DBLP:conf/icml/ChenW00T0CYW23}, CAV-MAE~\cite{DBLP:conf/iclr/GongRLHKKG23} for audio). The extracted Video Features and Audio Features are then fed into two models: the Multimodal Model combines both features for joint analysis, while the Unimodal Model uses only video features to evaluate the impact of audio information. Both models utilize ActionFormer~\cite{DBLP:conf/eccv/ZhangWL22} for temporal action localisation. The outputs from these models are then WBF~\cite{DBLP:journals/ivc/SolovyevWG21}to produce the final action localisation results. This fusion leverages the strengths of both models, enhancing the accuracy and reliability of the action detection.
\label{fig2}}
\end{figure*}

\section{Method}
In this section, we describe the methodology employed to address the Temporal Action Localisation (TAL) task. Our approach leverages state-of-the-art models for both video and audio feature extraction, followed by multimodal and unimodal model training, and fusion of predictions using WBF~\cite{DBLP:journals/ivc/SolovyevWG21}. The key components of our method include dataset augmentation, feature extraction, model training, and prediction fusion.
An overview of the method is illustrated in Figure \ref{fig2}.

\subsection{Dataset}
One of the main challenges in TAL tasks is the availability of sufficient training data. To enhance the generalization of our models, we augmented the competition dataset by incorporating overlapping labels from the Something-SomethingV2 dataset. Something-SomethingV2 is a large-scale, crowd-sourced video dataset focusing on basic interactions with everyday objects. We identified classes that overlap with the competition dataset and extracted corresponding samples to supplement our training set. This augmentation expanded the diversity of the training data and helped our models better capture temporal dependencies in various actions. 

\subsection{Video Feature Extraction}
\textbf{UMT.} Unified Multimodal Transformers (UMT)~\cite{DBLP:conf/cvpr/LiuLWCSQ22} is designed to address the dual tasks of video moment retrieval and highlight detection by leveraging multimodal learning and model flexibility. Unlike traditional methods that only consider visual data, UMT~\cite{DBLP:conf/cvpr/LiuLWCSQ22} incorporates video, audio, and textual information, making it capable of handling different combinations and reliability levels of these modalities. For example, when textual data is unavailable or unreliable, UMT~\cite{DBLP:conf/cvpr/LiuLWCSQ22} can still perform effectively by focusing on the remaining modalities. This flexibility allows UMT~\cite{DBLP:conf/cvpr/LiuLWCSQ22} to adapt to natural variations in the data without requiring multiple specialized models. UMT~\cite{DBLP:conf/cvpr/LiuLWCSQ22} has demonstrated its effectiveness on several benchmark datasets, outperforming state-of-the-art approaches in both video moment retrieval and highlight detection tasks.

\textbf{VideoMAEv2.} VideoMAEv2~\cite{DBLP:conf/cvpr/WangHZTHWWQ23} is an enhanced version of VideoMAE~\cite{DBLP:conf/nips/TongS0022}, designed to efficiently pre-train large-scale video models using a dual masking strategy. This approach involves applying separate masks to both the encoder and the decoder, reducing computational costs while maintaining strong performance. The encoder processes a small subset of visible tokens, while the decoder operates on a combination of latent features and additional masked tokens. The final loss is computed based on the reconstruction of masked pixels:

\begin{equation}
\ell = \frac{1}{(1 - \rho_d)N} \sum_{i \in M_d \cap M_e} \left| I_i - \hat{I}_i \right|^2,
\end{equation}

\noindent where $M_e$ and $M_d$ are the masking maps for the encoder and decoder, respectively, and $\hat{I}_i$ represents the reconstructed pixels. This dual masking approach improves training efficiency, enabling VideoMAEv2~\cite{DBLP:conf/cvpr/WangHZTHWWQ23} to scale to billion-parameter models.

Additionally, VideoMAEv2~\cite{DBLP:conf/cvpr/WangHZTHWWQ23} employs a progressive training pipeline. It starts with unsupervised pre-training on a large-scale, unlabeled hybrid dataset, followed by supervised post-pre-training on a labeled dataset to incorporate semantic knowledge, and finally fine-tuning on specific target tasks. This design allows VideoMAEv2~\cite{DBLP:conf/cvpr/WangHZTHWWQ23} to achieve state-of-the-art results in video action recognition and temporal action detection, pushing the performance limits of large video transformers.

\subsection{Audio Feature Extraction}

\textbf{BEATs.} BEATs~\cite{DBLP:conf/icml/ChenW00T0CYW23} is an iterative self-supervised learning framework for audio. It introduces an acoustic tokenizer that generates discrete, semantic-rich labels for general audio pre-training. Through iterative training, the tokenizer and the audio model improve each other. BEATs~\cite{DBLP:conf/icml/ChenW00T0CYW23} utilizes a mask and label prediction strategy, enabling it to capture high-level semantic information in audio. The model achieves state-of-the-art performance across various audio classification benchmarks, including AudioSet-2M and ESC-50, surpassing previous models in accuracy and efficiency.

\textbf{CAV-MAE.} CAV-MAE~\cite{DBLP:conf/iclr/GongRLHKKG23} extends the traditional Masked Autoencoder (MAE) framework to a multimodal setting, integrating both contrastive learning and masked data modeling. By jointly encoding audio and visual information, CAV-MAE~\cite{DBLP:conf/iclr/GongRLHKKG23} learns coordinated audio-visual representations. Contrastive learning is applied between the two modalities, while masked data modeling is used for reconstructing the missing portions of the inputs. The total loss is computed as:
\begin{equation}
L_{\text{CAV-MAE}} = L_r + \lambda_c \cdot L_c,
\end{equation}
where $L_r$ is the reconstruction loss and $L_c$ is the contrastive loss. This combination allows CAV-MAE to outperform previous models on both audio-visual classification and retrieval tasks, achieving state-of-the-art results on datasets like VGGSound and AudioSet.

\subsection{Prediction Fusion using WBF}
We trained two versions of the model:

\textbf{Multimodal Model}: This model utilizes both video and audio features to predict the start and end times of actions, as well as classify their types. The combination of modalities enhances the model's ability to understand complex interactions within the video.

\textbf{Unimodal Model}: In this version, only video features are used for action localisation. This allows us to evaluate the performance of visual features independently of audio inputs.

After training both multimodal and unimodal models, we combined their predictions using Weighted Box Fusion (WBF)~\cite{DBLP:journals/ivc/SolovyevWG21}. WBF~\cite{DBLP:journals/ivc/SolovyevWG21} is a method for merging predictions from multiple models by averaging the bounding box coordinates and confidence scores, weighted by their respective accuracies. This technique allows us to leverage the strengths of both the multimodal and unimodal models, resulting in more accurate and reliable action localisation.

\begin{figure}[ht]
    \centering
    \includegraphics[width=\linewidth]{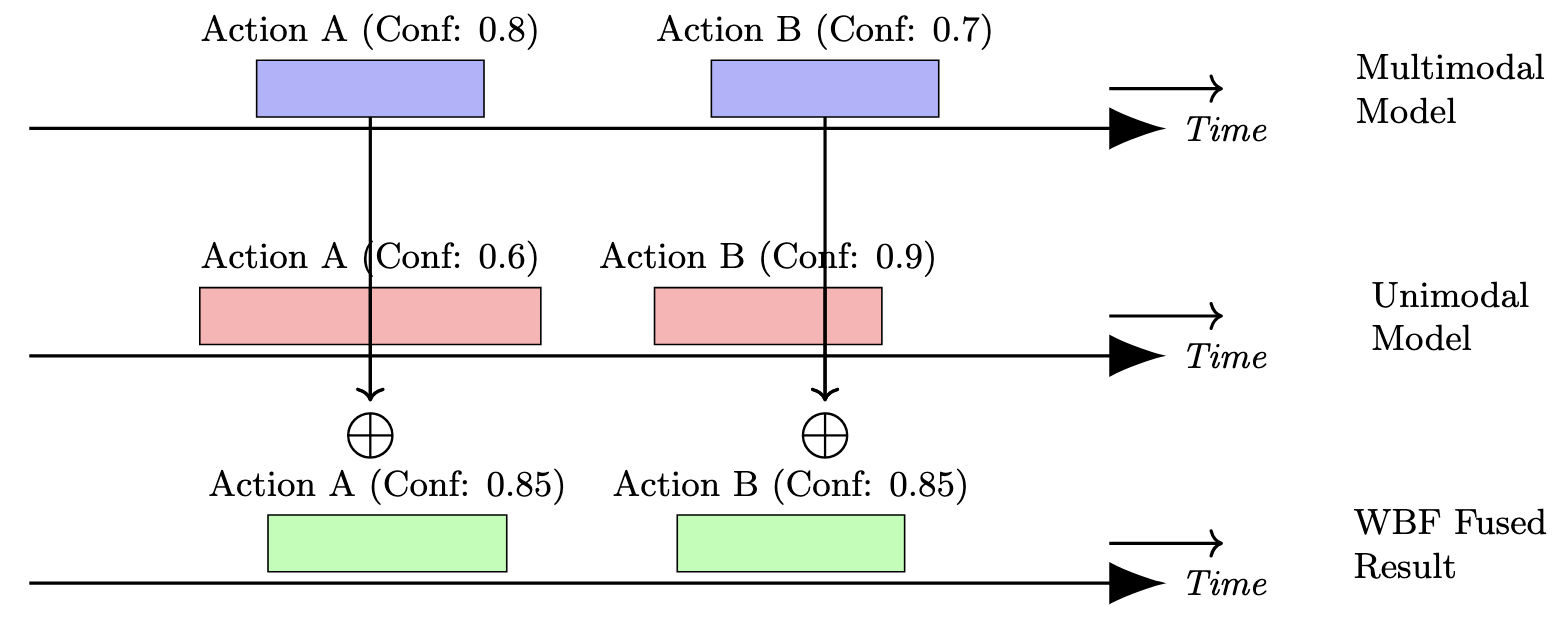}
    \caption{An illustration of the Weighted Box Fusion process in Temporal Action Localisation. Predictions from the multimodal and unimodal models are fused to produce a more accurate action localisation result.}
    \label{fig:wbf_tal}
\end{figure}

\section{Experiments}

\textbf{Datasets.} We evaluated our method on the Perception Test Temporal Action Localisation dataset, which consists of high-resolution videos with multiple annotated action segments. To augment the training data, we incorporated overlapping action classes from the Something-SomethingV2 dataset, enriching the diversity and quantity of training samples.

\textbf{Metric.} The evaluation metric is mean Average Precision (mAP). It measures the average precision across all classes and IoU thresholds. The mAP is calculated by evaluating the precision and recall at various IoU thresholds between the predicted action segments and the ground truth annotations.

\textbf{Comparative Experiments.} We compared our method with several state-of-the-art models:

\begin{table}[h]
\centering
\caption{Comparison with State-of-the-Art Models}
\label{tab:comparison}
\begin{tabular}{lcccc}
\toprule
\textbf{Model} & \textbf{Avg mAP} \\
\midrule
Baseline Model                & 16.0 \\
UMT~\cite{DBLP:conf/cvpr/LiuLWCSQ22}         & 47.3 \\
VideoMAEv2~\cite{DBLP:conf/cvpr/WangHZTHWWQ23}  & 49.1 \\
\textbf{Ours (Multimodal)}    & \textbf{53.2} \\
\bottomrule
\end{tabular}
\end{table}

Table \ref{tab:comparison} shows that our proposed method outperforms existing models across all IoU thresholds.

\textbf{Ablation Studies.} To assess the contribution of each component, we conducted ablation studies on Table \ref{tab:ablation}. The results in Table \ref{tab:ablation} indicate that each component contributes to the overall performance improvement.

\begin{table}[h]
\centering
\caption{Ablation Study on Method Components}
\label{tab:ablation}
\begin{tabular}{lcccc}
\toprule
\textbf{Method Variants}& \textbf{Avg mAP} \\
\midrule
+ Audio Features     & 49.5 \\
+ Combine Different Video Features   & 51.2 \\
+ Augement Dataset          & 53.2 \\
+ WBF                             & 54.9 \\
\bottomrule
\end{tabular}
\end{table}

\section{Conclusion}
This report presents a comprehensive evaluation of temporal action localisation (TAL) methods, highlighting the selection and application of advanced models such as UMT~\cite{DBLP:conf/cvpr/LiuLWCSQ22} and VideoMAEv2~\cite{DBLP:conf/cvpr/WangHZTHWWQ23} as the core video feature extraction algorithms, and BEATs~\cite{DBLP:conf/icml/ChenW00T0CYW23}, CAV-MAE~\cite{DBLP:conf/iclr/GongRLHKKG23} for audio feature extraction.. By leveraging diverse datasets, including the Perception Test dataset augmented with overlapping labels from Something-SomethingV2, and employing advanced techniques such as the incorporation of both multimodal and unimodal inputs and prediction fusion using WBF~\cite{DBLP:journals/ivc/SolovyevWG21}, the method demonstrated significant improvements in action localisation performance. Despite the limited impact of certain methods, the overall approach effectively addressed the challenges of temporal action localisation, leading to a top ranking in the final test phase.

{\small
\bibliographystyle{unsrt}
\bibliography{main}
}
\end{document}